\documentclass{article}
\usepackage{ijcai26}

\usepackage{times}
\usepackage{multirow}
\usepackage{array}     
\usepackage{soul}
\usepackage{xcolor}
\usepackage{url}
\usepackage{xcolor}
\usepackage[hidelinks]{hyperref}
\usepackage[utf8]{inputenc}
\usepackage[small]{caption}
\usepackage{graphicx}
\usepackage{amsmath}
\usepackage{amsthm}
\usepackage{booktabs}
\usepackage{algorithm}
\usepackage{algorithmic}
\usepackage{dblfloatfix} 
\usepackage{amssymb} 

\usepackage{placeins} 
\usepackage[switch]{lineno}
\title{

Energy-Regularized Spatial Masking: A Novel Approach to Enhancing Robustness and Interpretability in Vision Models\\
}

\author{
Tom Devynck$^{1}$ \and
Djamel Bouchaffra$^{1}$ \and
Nadjib Lazaar$^{3}$ \and
Mustapha Lebbah$^{1}$\\
Bilal Faye$^{2}$ \and
Hanane Azzag$^{2}$
\affiliations
$^{1}$ DAVID Lab, UVSQ, Paris-Saclay University, Versailles, France\\
$^{2}$ LIPN, UMR CNRS 7030, Sorbonne Paris Nord University, Villetaneuse, France\\
$^{3}$ LISN, Paris-Saclay University, Saclay, France\\
\emails
tom.devynck@uvsq.fr,
djamel.bouchaffra@gmail.com,\\
lazaar@lisn.fr,
mustapha.lebbah@uvsq.fr
faye@lipn.univ-paris13.fr,
azzag@lipn.univ-paris13.fr}

\begin{document}

\maketitle

\begin{abstract}
Deep convolutional neural networks achieve remarkable performance by exhaustively processing dense spatial feature maps, yet this brute-force strategy introduces significant computational redundancy and encourages reliance on spurious background correlations. As a result, modern vision models remain brittle and difficult to interpret. We propose Energy-Regularized Spatial Masking (ERSM), a novel framework that reformulates feature selection as a differentiable energy minimization problem. By embedding a lightweight Energy-Mask Layer inside standard convolutional backbones, each visual token is assigned a scalar energy composed of two competing forces: an intrinsic Unary importance cost and a Pairwise spatial coherence penalty. Unlike prior pruning methods that enforce rigid sparsity budgets or rely on heuristic importance scores, ERSM allows the network to autonomously discover an optimal information-density equilibrium tailored to each input. We validate ERSM on convolutional architectures and demonstrate that it produces emergent sparsity, improved robustness to structured occlusion, and highly interpretable spatial masks, while preserving classification accuracy. Furthermore, we show that the learned energy ranking significantly outperforms magnitude-based pruning in deletion-based robustness tests, revealing ERSM as an intrinsic denoising mechanism that isolates semantic object regions without pixel-level supervision. Code is available at \url{https://github.com/Tom-Dvk/ERSM}
\end{abstract}

\section{Introduction}

The success of deep learning in visual recognition has been driven 
by dense architectures that process all spatial locations uniformly. 
While effective, this exhaustive computation introduces significant 
redundancy and encourages models to rely on spurious background correlations 
rather than learning semantically meaningful structure. 
This limitation is especially pronounced in fine-grained classification, 
where models frequently overfit to contextual cues instead of object-specific features.
To address this issue, recent work has explored dynamic inference, 
enabling input-adaptive computation. Methods such as 
DynamicViT~\cite{rao2021dynamicvit}, EViT~\cite{liang2022evit}, 
and ToMe~\cite{bolya2023tome} demonstrate the benefits of adaptivity 
through token pruning or merging. However, these approaches typically 
depend on heuristic importance estimates and predefined sparsity schedules, 
which can discard subtle semantic information in favor of computational efficiency.


\begin{figure}
    \centering
    \includegraphics[width=1\linewidth]{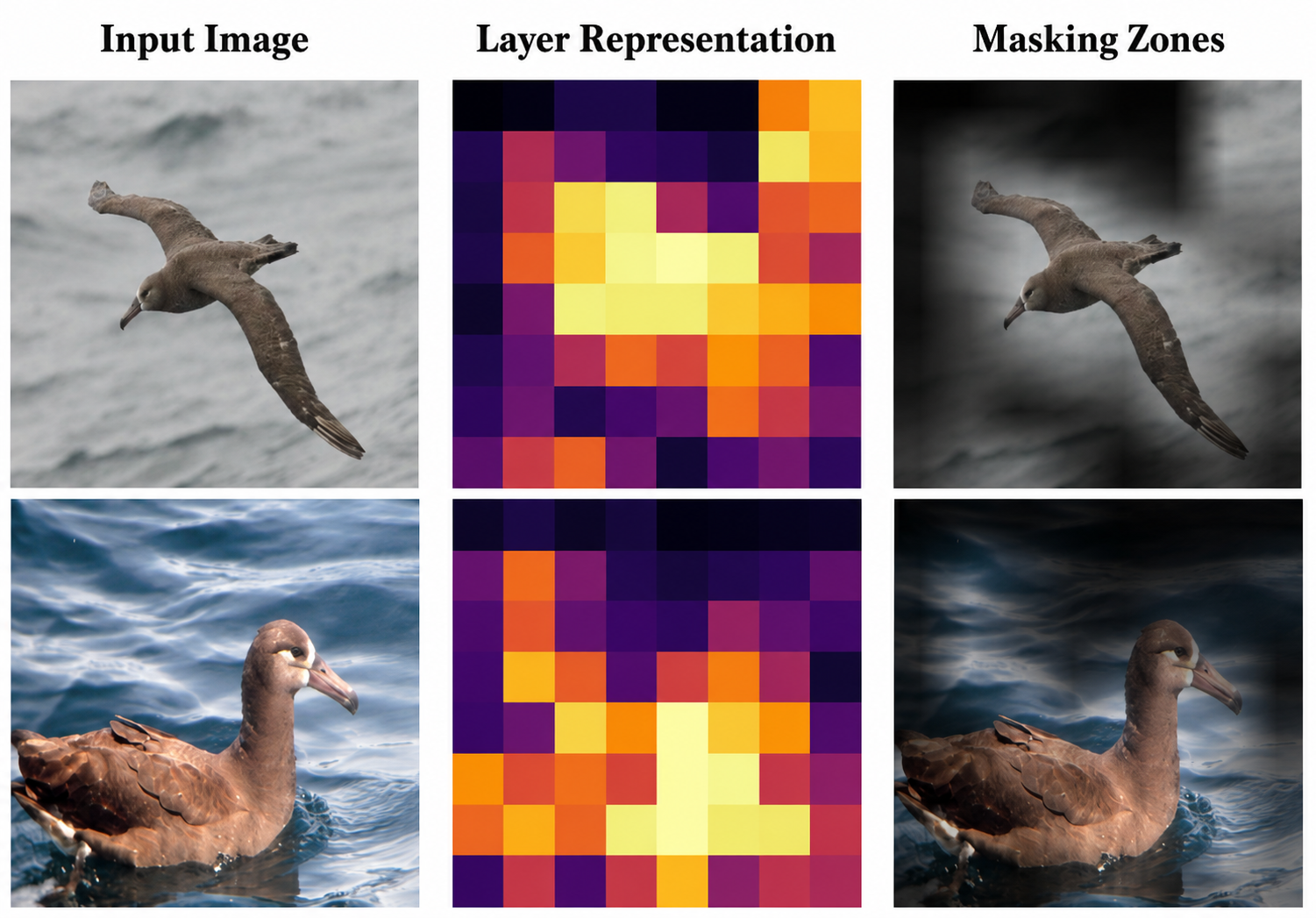}
    \caption{Energy-Based Information Preservation. ERSM adaptively retains semantically relevant features while suppressing background noise.} 
    \label{fig:introducing}
\end{figure}

Beyond efficiency, interpretability has become a central concern in modern vision
systems. 
Prototype-based models such as ProtoPNet and its deformable 
extension~\cite{Donnelly2022DeformableProtoPNet} provide human-interpretable 
explanations by matching latent image regions to learned visual prototypes, 
explicitly modeling geometric variability. 
However, these methods remain inherently similarity-driven: 
relevance is inferred from proximity to stored exemplars rather than from an 
intrinsic principle of information preservation. 
As a result, they do not directly address a more fundamental question: how much of the visual signal is truly required for correct reasoning?


In contrast, we argue that efficient and interpretable visual representations 
emerge naturally from a principle of energy minimization. 
Inspired by statistical mechanics and recent connections between game theory 
and deep learning~\cite{BouchaffraYFLA25}, we introduce Energy-Regularized 
Spatial Masking (ERSM), a physics-inspired framework that embeds a differentiable 
energy system within standard convolutional backbones.
Our first contribution is to adapt energy-based modeling from global image 
generation to internal feature selection, allowing spatial relevance to be 
inferred from an intrinsic information-preservation principle rather than 
heuristic importance scores. 
Second, we model spatial tokens as an interacting system whose forward pass 
corresponds to an implicit energy-minimization trajectory, eliminating the 
need for predefined sparsity budgets and enabling input-adaptive computation. 
Third, we demonstrate that this energy regularization induces emergent, 
contiguous, and semantically aligned spatial masks that improve robustness 
and interpretability without segmentation supervision, while remaining 
lightweight and fully compatible with conventional convolutional architectures.

\section{Related Work}

A large body of work seeks to reduce spatial redundancy to lower computational costs.
DynamicViT~\cite{rao2021dynamicvit} enforces fixed token keep-rates via a Gumbel-Softmax predictor, 
EViT~\cite{liang2022evit} prunes tokens using class-attention scores, 
and ToMe~\cite{bolya2023tome} further merges similar tokens. 
While effective for inference speed, these methods treat redundancy purely as a 
computational concern. 
In contrast, our ERSM considers redundancy a representational liability: 
instead of physically removing tensors for wall-clock gains, 
it masks features to maximize the signal-to-noise ratio. 
This approach preserves the dense tensor structure of standard CNNs such as ResNet, 
while conferring the robustness and interpretability benefits typically 
associated with sparse representations.


Energy-Based Models (EBMs) have regained attention for modeling complex data 
distributions~\cite{lecun2006tutorial,grathwohl2020jem} and improving 
out-of-distribution detection. 
However, conventional EBMs assign energy at the global input level, 
primarily for generative modeling. 
In contrast, our ERSM applies EBM principles to internal feature selection, 
interpreting the forward pass as an energy-minimization trajectory. 
This perspective bridges generative, physics-inspired principles with 
discriminative representation learning, enabling the network to 
selectively preserve semantically relevant features.


While post-hoc methods like Grad-CAM~\cite{selvaraju2017gradcam} 
visualize importance, they do not influence the model’s reasoning. 
Learnable masking approaches, such as Double-Win Quant~\cite{yang2022doublewin}, 
integrate selection during training but often produce scattered, 
noisy activation maps due to a lack of spatial constraints. 
To address this, we draw inspiration from classical computer vision, 
where Graph-Cut–based energy minimization enforces spatial 
continuity~\cite{5654046}. 
Our ERSM adapts this principle to deep feature space: by introducing an explicit pairwise interaction term, we enforce spatial coherence as a core modeling principle, yielding compact, object-aligned masks that are more interpretable than unstructured gating mechanisms.



 \section{Method}

We propose Energy-Regularized Spatial Masking (ERSM), 
a framework that reformulates feature pruning as a 
physics-inspired energy minimization problem. Unlike 
heuristic approaches that impose hard sparsity constraints, 
ERSM allows the network to autonomously discover an 
input-adaptive sparsity pattern. We introduce a differentiable 
Energy-Mask Layer in which retaining a visual token incurs 
an explicit energy cost. During training, the network balances 
task performance against the energetic cost of processing 
redundant spatial features, converging toward an equilibrium 
that maximizes information density. Notably, ERSM does not seek 
the exact equilibrium of classical Energy-Based Models, 
as the energy function does not fully encode all 
particle interactions; rather, it provides a tractable, 
discriminative approximation tailored to representation learning.


\subsection{Feature Representation}
To operationalize the energy formulation, we 
first discretize the continuous spatial feature 
map into a set of decision units. Given an 
intermediate feature map $\mathcal{F} 
\in \mathbb{R}^{C \times H \times W}$ produced 
by the convolutional backbone, we apply a spatial 
unfolding operation that partitions the grid into 
non-overlapping $d \times d$ patches. 
This yields a sequence of $N = HW / d^{2}$ 
tokens, each encoding the local texture and 
semantic content of its corresponding region.
Let $\mathbf{P} \in \mathbb{R}^{N \times D}$ denote the 
resulting token matrix, where $D = C \cdot d^{2}$ is the 
flattened feature dimension. To stabilize energy 
computation across varying feature magnitudes, 
each token $\mathbf{p}_i \in \mathbb{R}^{D}$ is 
$L_2$-normalized, 
yielding 
$\hat{\mathbf{p}}_i = \mathbf{p}_i / \lVert \mathbf{p}_i \rVert_2$.



\subsection{Energy Formulation}

We define the energy $E_i$ of a token as the cost of 
retaining it in the active set. Under the principle of 
energy minimization, the network is encouraged to 
suppress tokens with high retention cost. 
For a feature token $\mathbf{p}_i \in \mathbb{R}^{D}$, 
the total energy $E_i$ is defined as:

\begin{equation}
\label{eq:Ei}
\begin{split}
    E_i = & \lambda_{unary} \times \mathrm{softplus} \underbrace{\left(\mathbf{W}_{\mathrm{mask}}^\top \hat{\mathbf{p}}_i + b_i\right)}_{\text{$E^{\text{unary}}_{i}$: Unary Potential}}  \\
    & +\lambda_{pair} \times \mathrm{softplus}\underbrace{\left (\sum_{j\in\mathcal{N}(i)} \hat{\mathbf{p}}_i^\top \hat{\mathbf{p}}_j\right)}_{\text{$E^{\text{pair}}_{i}$: Pairwise Potential}} \\
    & =  E_i^{u+} + E_i^{p+} 
\end{split}
\end{equation}





The first term, $E_i^{u+}$, corresponds to the unary 
potential $E_i^{\text{Unary}}$, passed through a 
$\mathrm{softplus}$ function to enforce positivity 
and scaled by $\lambda_{\text{unary}}$. 
The vector 
$\mathbf{W}_{\mathrm{mask}} \in \mathbb{R}^{D}$ is a 
learned noise template that captures dominant directions 
of background redundancy rather than object-specific features. 
Tokens that align strongly with $\mathbf{W}_{\mathrm{mask}}$
incur a large positive unary energy, increasing their 
retention cost. In contrast to attention mechanisms 
that promote aligned features, this alignment penalizes 
tokens resembling background patterns. Tokens that are 
semantically distinct (i.e., weakly aligned with the 
noise template) incur low energy and are therefore 
inexpensive to retain, allowing informative content to 
be preserved.


The second term, $E_i^{p+}$, corresponds to the 
pairwise potential $E_i^{\text{pair}}$, passed through 
a $\mathrm{softplus}$ function to enforce positivity 
and scaled by $\lambda_{\text{pair}}$. 
This term explicitly penalizes spatial redundancy, 
encouraging coherent spatial selection.
We define $\mathcal{N}(i)$ as the Moore (8-connected)
neighborhood of token $i$ on the token grid. 
The pairwise term aggregates the cosine similarity 
between the central token $\mathbf{p}_i$ and its 
spatial neighbors. In visually homogeneous regions, 
neighbor similarity is high, yielding a large 
positive energy penalty and increasing the cost of 
retaining redundant feature clusters. Minimizing 
this potential encourages the model to suppress 
spatially redundant tokens while preserving 
distinctive, low-cost signals.


\begin{algorithm}[h!]
\caption{Energy Mask (Unary and Full Variants) }
\label{alg:energymask}
\begin{algorithmic}[1]

\REQUIRE Feature map 
$X \in \mathbb{R}^{C \times H \times W}$, 
patch size $D$, weights $\lambda_{unary},\lambda_{pair}$
\ENSURE Masked feature map $\tilde{X}$ and diagnostics 
$(m,z,E^{u+},E^{p+})$
\STATE
$\{\mathbf p_i\}_{i=1}^{N} \leftarrow \text{Unfold}(X;D,D)$
\hfill\textit{// extract $N=HW / D^2$ spatial patches}

\STATE $\hat{\mathbf{p}}_i = \mathbf{p}_i / (\lVert \mathbf{p}_i \rVert_2+ \varepsilon)$ \quad $\forall i$  
\hfill\textit{// patch normalization}

\STATE 
$z_i=\mathbf{w}^\top \hat{\mathbf{p}}_i + b_i$
\quad $\forall i$  
\hfill\textit{// learned patch importance score}

\STATE $m_i \leftarrow \sigma(-z_i)$ \quad $\forall i$  
\hfill\textit{// soft keep probability}

\STATE $E^{u+}_i \leftarrow \lambda_{unary} \cdot \text{softplus}(z_i)$ \quad $\forall i$
\hfill\textit{// unary energy cost}

\IF{$\lambda_{pair} > 0$}
    \STATE 
     $E_i^{pair} \leftarrow \sum_{j \in \mathcal{N}_8(i)} \langle \hat{\mathbf p}_i, \hat{\mathbf p}_j \rangle$ \quad $\forall i$  
    \hfill\textit{// local feature coherence: Moore neighborhood (8-connected)}
    
    \STATE 
    $E^{p+}_i \leftarrow \lambda_{pair} \cdot \text{softplus}(E_i^{pair})$ 
    \quad $\forall i$  
    \hfill\textit{// pairwise energy cost}
\ELSE
    \STATE 
    $E^{p+}_i \leftarrow 0$
    \quad $\forall i$
\ENDIF

\STATE 
$\tilde{\mathbf p}_i \leftarrow m_i \cdot \mathbf p_i$ \quad $\forall i$ \hfill\textit{// soft patch suppression}

\STATE 
$\tilde{X} \leftarrow \text{Fold}(\{\tilde{\mathbf p}_i\}; H, W, D, D)$  
\hfill\textit{// reconstruct masked feature map}

\RETURN $\tilde{X}$ and $(m,z,E^{u+},E^{p+})$

\end{algorithmic}
\end{algorithm}

\begin{figure*}[t]
    \centering
    \includegraphics[width=1\linewidth, height=5cm]{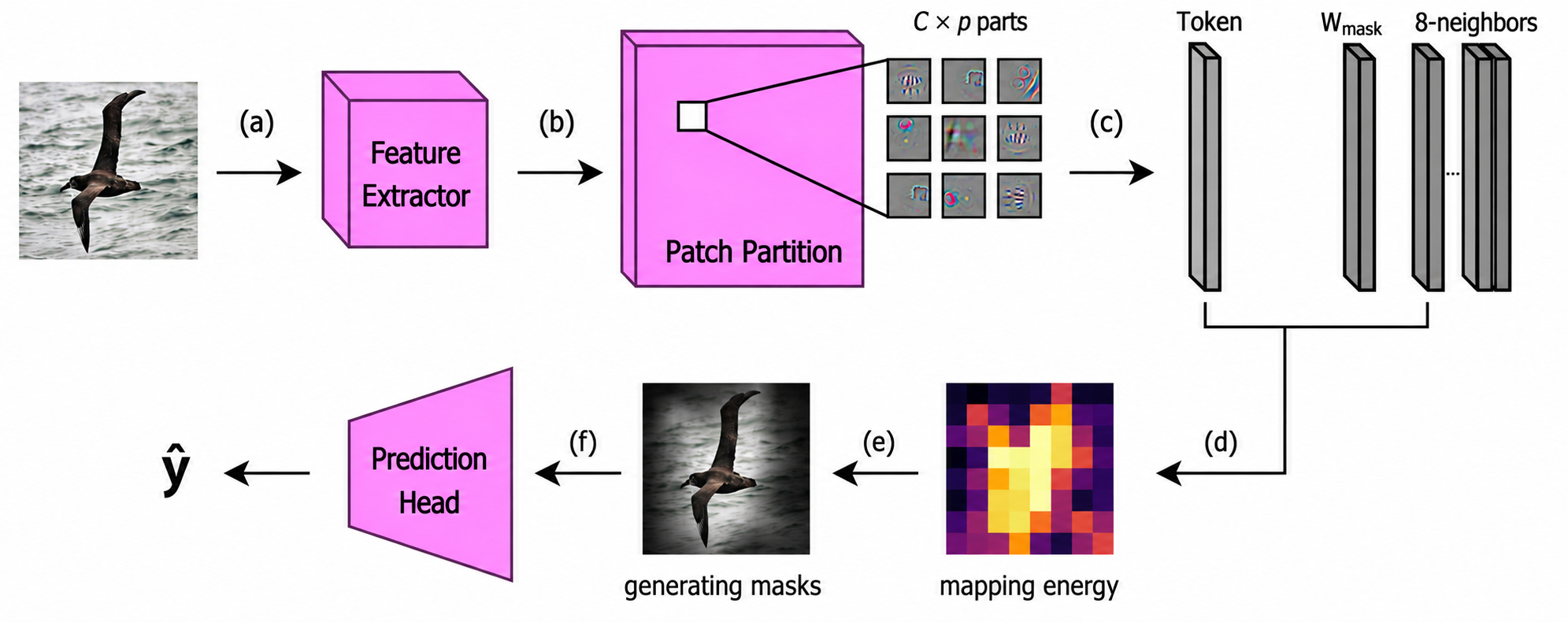}
 
 \caption{Overview of the Energy-Regularized Spatial Masking (ERSM) framework. (a) A frozen feature extractor encodes the input image. (b–c) The feature map is partitioned into latent tokens. (d) Unary and 8-neighbor pairwise interactions define a token-wise energy. (e) The resulting spatial energy map suppresses redundant background regions. (f) The refined features are passed to the prediction head for classification.}
    \label{fig:placeholder}
\end{figure*}
\subsection{Differentiable Gating Policy}
To enable end-to-end optimization, we relax the discrete token selection problem into a continuous, differentiable formulation. Unlike standard attention mechanisms that learn to attend based solely on task performance, our framework learns a gating policy guided by energy minimization.

\paragraph{Training Objective: Emergent Sparsity.}
A key advantage of our approach is that, 
unlike methods enforcing a fixed sparsity budget 
$\rho$ (e.g., retaining exactly $k$ patches), 
we impose a cost on feature retention. The network 
can retain as many tokens as needed, as long as their
contribution to classification performance justifies 
their energy cost. The regularization loss is 
defined as the expected energy of the active system. 
For a batch of $N$ tokens, this corresponds to the 
mean energy of the retained set:
\begin{equation}
    \mathcal{L}_{\text{reg}} = \frac{1}{N} \sum_{i=1}^{N} m_i \cdot E_i
\end{equation}

where $m_i = \sigma(-E_i^{\text{unary}}) \in [0,1]$ 
is the token’s continuous retention probability, 
and $E_i$ its energy cost.



The total training objective combines the task-specific loss with the energy regularization term:

\begin{equation}
    \mathcal{L}_{\text{total}} = \mathcal{L}(\mathbf{y}, \hat{\mathbf{y}}) + \mathcal{L}_{\text{reg}}
\end{equation}

Minimizing this objective leads to an autonomous 
trade-off: gradient descent drives $m_i \to 1$ 
(retain) only if the reduction in classification 
loss $\mathcal{L}$ outweighs the token’s energy 
cost $E_i$. This produces emergent sparsity, 
where the number of retained tokens adapts 
dynamically to the semantic complexity of each 
input rather than being fixed by a hyperparameter.

\paragraph{Amortized Inference via Implicit Distillation.}
While the global energy $E_{\text{total}} = \sum_i E_i$
(including pairwise interactions) guides token 
selection during training, computing it at inference 
is costly. To address this, we employ amortized 
inference, where gating decisions rely solely on 
the local unary potential.
Token activation is interpreted as a stochastic 
equilibrium process. For each token $i$, 
the retention decision is driven by the unary penalty:

%
$$E^{\text{unary}}_{i} = \mathbf{W}_{\mathrm{mask}}^\top \hat{\mathbf{p}}_i + b_i.$$ 
where $\mathbf{W}_{\mathrm{mask}}$ serves as a 
learned redundancy prototype. Tokens aligned with 
this vector incur high energy and are likely 
suppressed. The probability of retaining a token 
is therefore:
\[
\sigma(-E^{\text{unary}}_{i}) = \frac{e^{-E^{\text{unary}}_{i}}}{1 + e^{-E^{\text{unary}}_{i}}}.
\]

The logic is consistent: low energy implies stability.
Tokens that poorly match the noise template 
(low $E_i^{\text{unary}}$) have 
$\sigma(-E_i^{\text{unary}}) \to 1$, while 
high-energy tokens are unstable and likely dropped. 
Although decisions are local, the weights 
$\mathbf{W}_{\text{mask}}$ are shaped by the 
full energy objective during training, 
effectively distilling global constraints into a 
fast, local inference policy.

While we refer to $E_{\text{total}} = \sum_i E_i$ 
as an energy function, it does not define an 
explicit energy-based probabilistic model; rather, 
it serves as a training-time regularizer inspired 
by energy-based principles.

\subsection{Gradient Dynamics}
Our framework decouples the decision policy from
energy evaluation, functioning as a differentiable 
mechanism embedded within the layer. The forward 
pass is efficient, producing gating decisions using 
only the local unary projection
$z_i = \mathbf{w}^\top \hat{\mathbf{p}}_i$,
while the backward pass evaluates decision quality 
by incorporating the costly pairwise interactions
into the loss. For a token with high spatial 
redundancy, the pairwise energy 
$E_i^{\text{pair}}$ is large. Minimizing the 
loss drives the retention mask $m_i$ toward zero.

Mechanically, this negative gradient pressure aligns the weights 
$\mathbf{w}$ more closely with $\hat{\mathbf{p}}_i$, increasing 
the noise projection $z_i$. 
Over training, the unary filter $\mathbf{w}$ 
effectively {\em memorizes} redundant textures, distilling global neighborhood 
constraints into a local linear projection. This enables future forward 
passes to suppress redundant features without explicitly computing pairwise 
interactions.



\section{Experiments}
We first validate ERSM in a controlled pilot setting before integrating it 
into standard deep learning backbones. 
Specifically, we evaluate its behavior 
and robustness on both shallow CNNs and larger architectures (e.g., ResNet 
variants~\cite{he2016resnet}), enabling us to isolate its effect independently 
of backbone depth and capacity. Importantly, these experiments are not intended 
to push state-of-the-art performance, but to cleanly quantify the contribution 
of the ERSM layer by comparing otherwise identical models with and without ERSM 
under controlled conditions.



\subsection{Controlled Proof of Concept}

In this study, ERSM is implemented as a lightweight \emph{Energy-Mask Layer} inserted after the final convolutional stage of a compact CNN trained on the Food-101 dataset~\cite{bossard2014food101}.
Input images are resized and center-cropped to $224 \times 224$. The backbone produces a feature map of size $C \times H \times W$ 
($256 \times 14 \times 14$ in our setting), which is partitioned 
into non-overlapping $d \times d$ patches ($d=2$), yielding a 
$7 \times 7$ grid of $N=49$ tokens. Each token is represented by 
a vector of dimension $D = C \cdot d^2$.

%
%
%
%
ERSM computes a per-token unary score $z_i=\mathbf{w}^\top \hat{\mathbf{p}}_i + b_i$ 
and retains tokens with probability $m_i=\sigma(-z_i)$. Training minimizes 
the classification loss augmented with the expected energy of retained tokens, 
$\mathcal{L}_{\text{total}}=\mathcal{L}_{\text{CE}}+\mathcal{L}_{\text{Reg}}$.
We perform a two-dimensional sensitivity analysis over the energy coefficients 
$\lambda_{\text{unary}}$ and $\lambda_{\text{pair}}$ using grid search. 
For each $(\lambda_{\text{unary}},\lambda_{\text{pair}})$ pair, the model is 
trained for 20 epochs on a 20\% subset of Food-101, and evaluated in terms of 
test accuracy and emergent sparsity. The configuration 
$\lambda_{\text{unary}}=\lambda_{\text{pair}}=10^{-3}$ consistently 
yields the best accuracy–sparsity trade-off and is used in all subsequent 
experiments.
\subsection{Results}
We compare the following variants: 
(i) a baseline CNN (four convolutional layers with ReLU and max-pooling) without masking; 
(ii) ERSM-Unary, using only the unary energy term ($\lambda_{\text{pair}}=0$); 
(iii) ERSM-Full, combining unary and pairwise energies ($\lambda_{\text{pair}}>0$); 
(iv) DropBlock~\cite{ghiasi2018dropblock}; 
(v) Spatial Dropout~\cite{tompson2015efficient}; and 
(vi) $L_0$ gating~\cite{louizos2018l0}, a patch-wise gating baseline with an $L_0$ sparsity penalty. All models are trained for 100 epochs using AdamW with cosine learning-rate scheduling.

ERSM layer is inserted after the final convolutional block, operating on $14\times14$ feature maps with $2\times2$ spatial patches and 256 channels.
All models are trained for 100 epochs with identical optimization settings. Table~\ref{tab:main_results} reports the peak test accuracy for each method.

\begin{table}[h!]
\centering
\setlength{\tabcolsep}{4pt}
\renewcommand{\arraystretch}{1}
\begin{tabular}{lcc}
\toprule
Model & Acc. (\%) $\uparrow$ & $\mathbb{E}[m_i]$ \\
\midrule
Baseline CNN       & 69.14 $\pm$ 0.22 & -- \\
DropBlock          & 69.53 $\pm$ 0.22 & --   \\
Spatial Dropout    & 68.88 $\pm$ 0.34 & --   \\
L0-Gating          & \textbf{69.86} $\pm$ 0.30 & 0.65 \\
\hline
ERSM-Unary (ours) & 69.76 $\pm$ \textbf{0.15} & 0.65 \\
ERSM-Full (ours)  & 69.73 $\pm$ 0.36 & 0.64 \\
\bottomrule
\end{tabular}
\caption{Classification performance on Food-101 (Mean $\pm$ Std over 5 seeds).}
\label{tab:main_results}
\end{table}

ERSM-Unary achieves performance comparable to the state-of-the-art L0-Gating ($69.76\%$ vs $69.86\%$), while learning a sparse soft gating over spatial patches, with an average keep probability of $\approx0.65$. 
This indicates consistent suppression of redundant 
spatial features while maintaining strong generalization. 
Notably, ERSM-Unary exhibits lower variance across random 
initializations than $L_0$ gating, suggesting more stable 
and robust convergence under the energy-based formulation.

%
This structured behavior emerges solely from the 
optimization objective, without explicit supervision or 
heuristic saliency mechanisms, highlighting ERSM’s 
ability to discover meaningful spatial organization 
within convolutional feature maps.


\begin{figure}[h!]
\centering
\includegraphics[width=7.5cm]{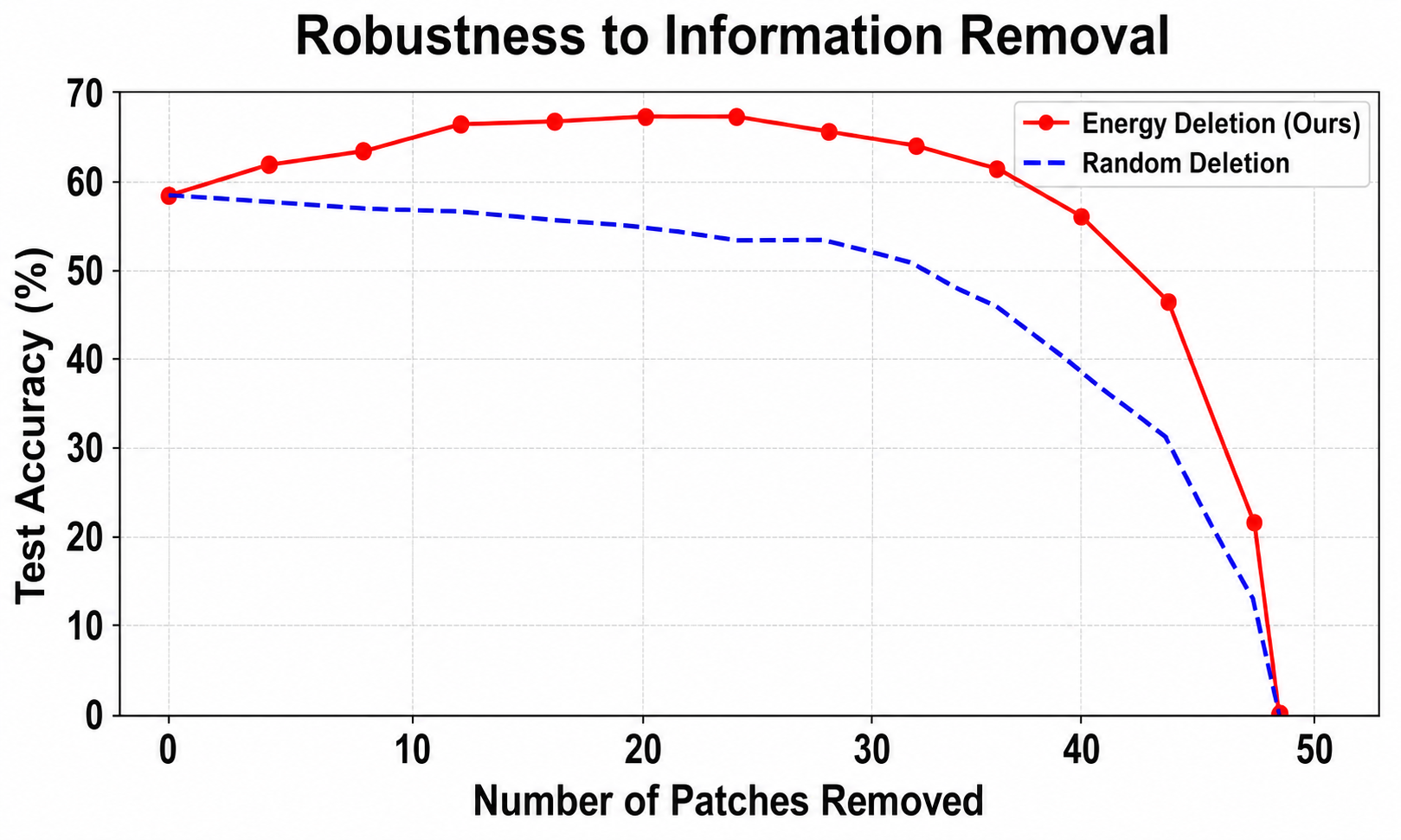}
\caption{Accuracy under progressive patch removal on the experimental CNN. ERSM preserves more accuracy than random deletion, peaking after removing 20\%--25\% of patches, highlighting its implicit feature-selection ability.}
\label{fig:robustness}
\end{figure}

To assess whether the learned energy scores capture meaningful spatial 
importance, we conduct a controlled deletion study on the test set. 
For each input, patches are ranked by their energy scores $z_i$, 
and the top-$k$ patches are progressively removed at the feature-map level. 
After deletion, global average pooling is rescaled by the ratio 
of total to remaining patches to preserve activation magnitude. 
This counterfactual evaluation differs from standard inference, 
which uses soft gating, but enables direct probing of the l
earned importance ranking.
As a baseline, random deletion is performed by sampling a random 
permutation of patch indices for each input and removing the first 
$k$ patches. The same hard suppression and pooling rescaling are 
applied in both cases, ensuring a fair comparison between energy-based 
and random deletion.


Figure~\ref{fig:robustness} shows classification accuracy as a function of the number of removed patches. Energy-guided deletion consistently preserves higher accuracy than random deletion across all removal levels. Interestingly, the ERSM curve initially increases, peaking after removing roughly 20\%–25\% of patches, before gradually declining. This suggests that the model identifies and discards not only redundant but also mildly harmful features, acting as an implicit feature-selection mechanism that improves generalization. ERSM thus learns a structured, semantically meaningful ordering of spatial features rather than relying on brittle or redundant activations.

\section{Scaling the Mechanism}
Having validated ERSM in a controlled setting, we next examine its 
scalability and generalization in deep, industry-standard architectures 
on complex real-world images.

\subsection{General Integration Strategy}

The core hypothesis of ERSM is that energy minimization provides a 
general principle for representation selection, independent of the 
underlying feature extractor. To test this, we evaluate ERSM as a 
modular, “plug-and-play” gating mechanism that can be inserted into 
standard convolutional backbones without architectural changes.
Although our experiments focus on convolutional backbones to isolate the 
contribution of ERSM independently of representational capacity, the 
Energy-Mask Layer is architecture-agnostic. ERSM operates on any sequence 
of spatial tokens, a condition satisfied by 
multiple vision paradigms. In Vision Transformers (ViT), tokens correspond 
directly to image patches, making ERSM integration structurally 
straightforward: unary and pairwise energies can be computed over patch 
embeddings prior to or between transformer blocks. This extends naturally 
to self-supervised variants such as MAE, which expose the same tokenized 
representation. Likewise, state-space vision architectures (e.g., Vim, VMamba) expose 
ordered spatial token sequences compatible with ERSM. We therefore view 
ERSM as a general energy-based token selection principle rather than a 
CNN-specific mechanism.
Scaling from small models to modern backbones introduces two challenges: 
(1) high-dimensional feature spaces with distributed semantic signals, 
and (2) maintaining robust energy rankings despite complex texture biases 
in real-world data. To isolate ERSM's ability to identify intrinsic features independently of backbone capacity, we adopt a frozen-backbone protocol: pretrained weights are fixed up to the insertion point, and only the ERSM parameters and final classifier are trained. This treats the backbone 
as a fixed semantic sensor and forces ERSM to act as a post-hoc attentional filter operating solely on existing feature correlations.

ERSM is inserted at the final feature layer of the backbone, just before pooling and classification. At this stage, representations encode high-level semantic parts rather than low-level textures, making energy-based ranking more reliable and task-relevant. Late insertion also minimizes computational cost by operating on the smallest feature map. Empirically, earlier insertion produced less stable energy scores and weaker gains, supporting the choice of a late-stage integration.

While the frozen-backbone protocol was deliberately chosen to isolate the 
effect of the ERSM layer independently of backbone capacity, ERSM remains 
fully differentiable and naturally compatible with end-to-end optimization. 
The pairwise energy term can be interpreted as a spatial smoothness prior, 
penalizing homogeneous redundant activations and encouraging the emergence 
of locally distinctive feature representations. In this setting, the 
backbone and ERSM layer would co-adapt during training, potentially 
yielding sparser and more semantically concentrated representations. We 
leave systematic investigation of end-to-end training dynamics as future 
work.


\section{Fine-Grained Benchmark Experiments}
To evaluate the generality of ERSM across semantic domains, 
we extend our experiments from generic classification to fine-grained 
visual recognition (FGVC). FGVC provides a challenging testbed for 
spatial masking, as correct recognition often relies on subtle, 
localized cues (e.g., facial patterns, fur texture, or vehicle part 
details) rather than global shape.
We evaluate ERSM on both generic and fine-grained benchmarks. 
For generic classification, we use Food-101~\cite{bossard2014food101}; 
for fine-grained tasks include Oxford-IIIT Pet~\cite{parkhi2012catsdogs} 
(37 classes) and CUB-200-2011~\cite{wah2011caltech} (200 bird species). 
ERSM is integrated into multiple ImageNet-pretrained backbones, 
including ResNet-50~\cite{he2016resnet}, ConvNeXt-Tiny~\cite{liu2022convnet}, 
and EfficientNetV2-S~\cite{tan2021efficientnetv2}. 
For each dataset, we report accuracy, masking, deletion robustness, and qualitative analyses.

\paragraph{Quantitative Results.}
Table~\ref{tab:detailed_results} summarizes the performance of 
ERSM-equipped backbones versus frozen baselines. We report test accuracy 
and total loss. All backbones are pretrained on ImageNet~\cite{deng2009imagenet} and fine-tuned for 20 epochs 
under the same settings used in the sensitivity analysis. We also 
explore the impact of input resolution and different patch sizes, 
while ensuring consistent tokenization.

\begin{table}[h]
\centering
\scriptsize
\renewcommand{\arraystretch}{0.95}
\setlength{\tabcolsep}{3pt}
\begin{tabular}{llcc cc cc}
\toprule
& & & & \multicolumn{2}{c}{\textbf{Baseline}} & \multicolumn{2}{c}{\textbf{ERSM}} \\
\cmidrule(lr){5-6}\cmidrule(lr){7-8}
\textbf{Dataset} & \textbf{Backbone} & \textbf{Res} & \textbf{P} & Acc & Loss & Acc & Loss \\
\midrule
\multirow{8}{*}{\textbf{CUB-200}}
 & ResNet-50         & 224 & 1 & 60.20 & 0.343 & 60.59 & 0.263 \\
 & ResNet-50         & 256 & 1 & 59.82 & 0.374 & 60.67 & 0.274 \\
 & ResNet-50         & 256 & 2 & 59.82 & 0.374 & 60.15 & 0.238 \\
 & ResNet-50         & 448 & 1 & 45.81 & 0.880 & 47.67 & 0.611 \\
 & ResNet-50         & 448 & 2 & 45.81 & 0.880 & 47.38 & 0.581 \\
 & ConvNeXt-Tiny     & 224 & 1 & 69.83 & 1.450 & 69.73 & 1.415 \\
 & ConvNeXt-Tiny     & 256 & 2 & 68.88 & 1.609 & 70.62 & 1.342 \\
 & EfficientNetV2-S  & 448 & 2 & 61.63 & 0.654 & 63.29 & 0.730 \\
\midrule
\multirow{2}{*}{\textbf{Food-101}}
 & ResNet-50         & 224 & 1 & 68.47 & 0.910 & 68.42 & 0.920 \\
 & ConvNeXt-Tiny     & 224 & 1 & 78.23 & 0.922 & 78.37 & 0.925 \\
\midrule
\multirow{2}{*}{\textbf{Oxford Pet}}
 & ResNet-50         & 224 & 1 & 91.14 & 0.098 & 89.45 & 0.134 \\
 & ConvNeXt-Tiny     & 256 & 1 & 93.32 & 0.215 & 93.21 & 0.967 \\
\midrule
\textbf{ImageNet-R}
 & ResNet-50         & 224 & 1 & 38.36 & 0.036 & 38.68 & 0.046 \\
\bottomrule
\end{tabular}
\caption{Baseline (frozen) vs.\ ERSM across resolutions (Res) and patch sizes (P). Configurations where P does not divide the feature map are omitted. Peak values shown.}
\label{tab:detailed_results}
\end{table}

Across all backbones and datasets, ERSM provides competitive performance compared to frozen baselines, particularly on CUB-200 and Food-101 where localized discriminative cues are critical. When no result is reported for a given patch size, this corresponds to cases where the chosen patch dimension does not divide the backbone feature map, preventing valid tokenization. Interestingly, ERSM does not uniformly improve performance 
across datasets. On Oxford-IIIT Pet, gains are marginal or 
slightly negative relative to the frozen baseline. We 
hypothesize that this reflects a reduced background-bias 
regime: pet images are typically object-centered with limited 
contextual ambiguity, leaving little opportunity for 
denoising mechanisms to improve representation quality. In 
contrast, datasets such as CUB-200 contain stronger 
contextual variability and localized discriminative cues, 
conditions under which ERSM consistently provides larger 
benefits. These observations suggest that ERSM is most 
effective when spurious correlations constitute a meaningful 
source of predictive bias.

\subsection{Ablation Study}
We perform a deeper analysis of ERSM dynamics using ResNet-50 on CUB-200 
with input size 256 and patch size 1, isolating the effects of spatial 
resolution and token granularity on energy minimization. The model is 
trained for 20 epochs with the dual objective of classification loss 
and energy regularization, converging stably to a peak test accuracy of 
$60.67\%$ and loss of $0.2737$. Importantly, it autonomously 
reaches a sparsity equilibrium with an average mask value of $0.60$, 
effectively suppressing roughly 40\% of the spatial field while 
preserving high classification performance.

\begin{figure}[h!]
    \centering
    \includegraphics[width=7.5cm]{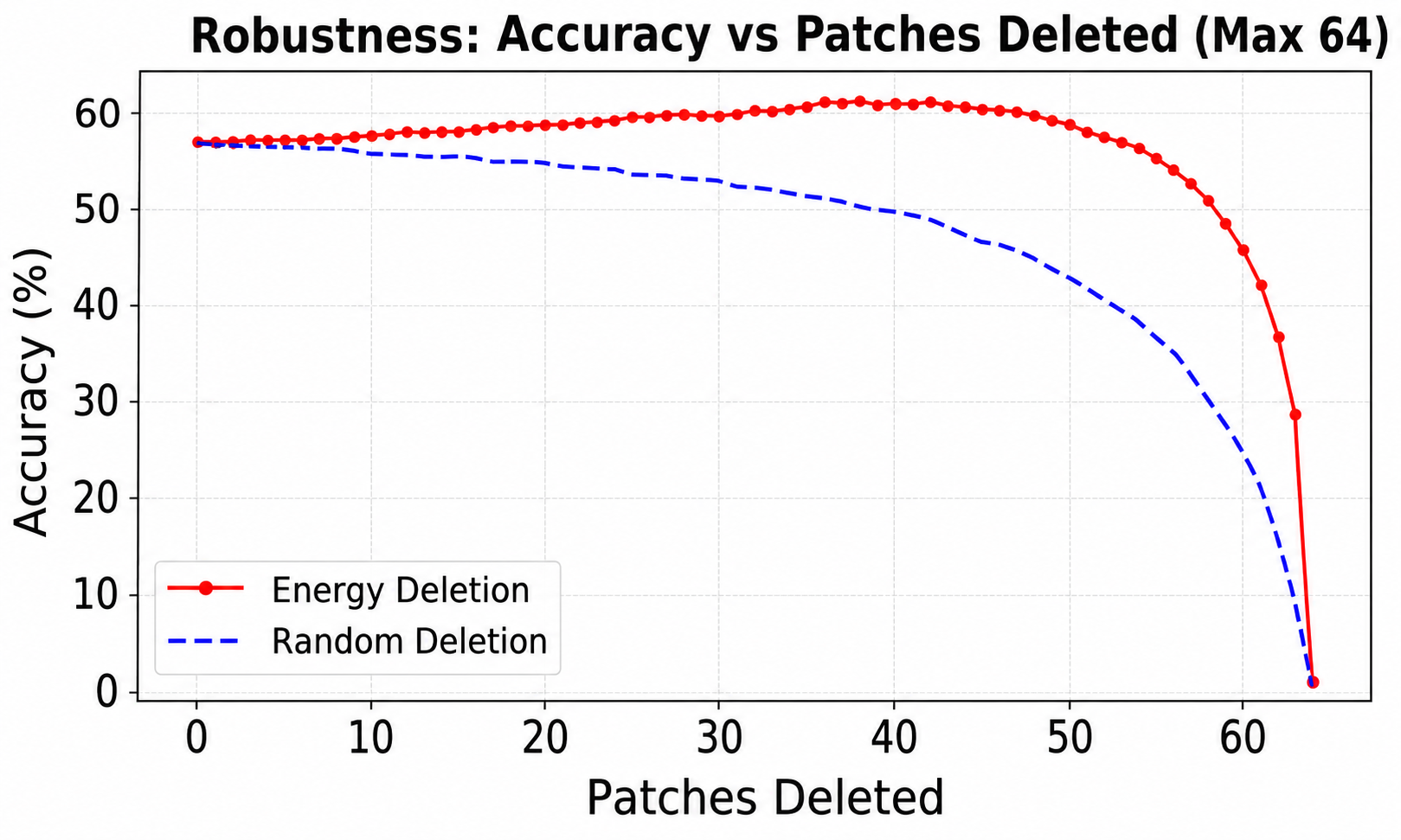}
    \caption{Robustness to Feature Removal on ResNet-50 Res 256 Patch 1. Comparison of classification accuracy under progressive patch deletion using Energy-based ranking (red) versus Random selection (blue).}
    \label{fig:robustness_resnet}
\end{figure}

To verify that discarded features are less informative, we analyze the 
robustness curve (Figure~\ref{fig:robustness_resnet}), which plots test 
accuracy as a function of the number of deleted patches, ranked by 
learned energy (highest first). This evaluation uses deterministic hard 
pruning, distinct from the soft gating in training. At each step $k$, 
the top-$k$ highest-energy patches are set to zero. 
To ensure accuracy degradation reflects information loss rather than 
reduced activation magnitude, we apply dynamic rescaling of the remaining 
features proportional to the retained token count, preventing the 
classifier from being biased by simple energy scaling.

The resulting robustness profiles in Figure~\ref{fig:robustness_resnet} 
reveal three key phenomena validating the effectiveness of energy-based ranking:
\textbf{(1)} Unlike the Random Deletion baseline, which shows immediate and
    monotonic accuracy loss, Energy-Guided Deletion exhibits a convex 
    trajectory. Removing the first 30–40 highest-energy patches actually 
    improves accuracy, indicating that the model assigns high energy to 
    background clutter and spurious correlations. Discarding these 
    distractor patches effectively denoises the representation before any 
    object-relevant features are affected;
\textbf{(2)} Performance remains well above baseline even after removing up to 
    50 of the 64 patches. The gap of up to 15 percentage points between 
    energy-guided and random deletion quantifies the strong semantic 
    alignment of the learned energy potential;
\textbf{(3)} Accuracy plateaus until a sharp “semantic cliff” occurs after 
    patch 55, where performance collapses. This indicates that the 
    essential discriminative features are concentrated in the final 
    10–15 lowest-energy tokens, while most of the feature map 
    contributes little or negatively to classification.


    
    

\subsection{Out-of-Distribution Robustness}

To evaluate whether ERSM learns transferable structural 
information rather than dataset-specific correlations, we 
perform a preliminary out-of-distribution evaluation 
using the ResNet-50 configuration (res = 224, p = 1) 
trained on a 200-class subset of ImageNet and evaluated on 
ImageNet-R. While top-1 accuracy remains comparable to the 
frozen baseline, ERSM preserves its characteristic deletion 
behavior under domain shift: energy-guided patch removal  degrades performance substantially more slowly than random 
deletion, indicating that the learned energy ranking captures structural semantic relevance rather than memorized 
in-distribution statistics. 

\subsection{Qualitative analysis}

\begin{figure}[h!]
    \centering
    \includegraphics[width=5.5cm]{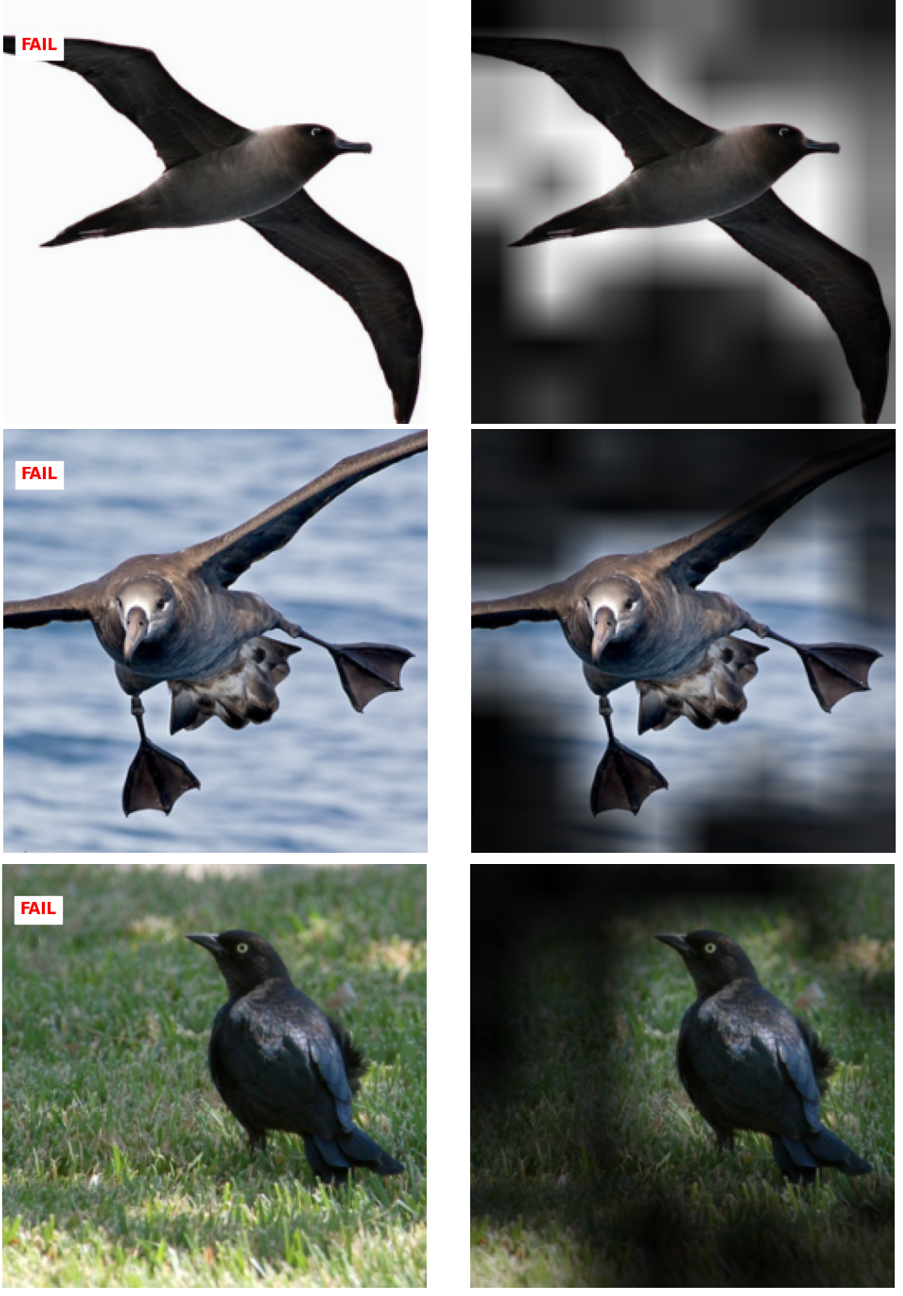}


    \caption{ERSM Improvements. Left: The original image with baseline prediction (Wrong). Right: The ERSM prediction (Correct), showing the learned mask fading out the misleading background.}
    \label{fig:improvements}
\end{figure}

\noindent To understand the decision-making of the ERSM layer, we examine visual 
examples from inference, categorizing test samples into 
\textit{Improvements} (where ERSM corrects baseline errors) and 
\textit{Failure Modes} (where ERSM misclassifies), providing a 
detailed view of model behavior.
Figure~\ref{fig:improvements} shows cases where the frozen ResNet-50 fails 
but the ERSM-augmented model predicts correctly. These examples often 
illustrate the “context trap” where the baseline relies on spurious 
background correlations. ERSM mitigates this by effectively masking 
misleading context, forcing the classifier to rely 
solely on intrinsic object features and thereby correcting predictions.
Notably, the faded visualizations are not post-hoc saliency maps but 
direct projections of the model’s active decision bottleneck. Token-level 
keep probabilities $\mathbf{m} \in [0,1]^{H \times W}$ from the ERSM layer 
are upsampled via bilinear interpolation and applied multiplicatively to 
the input image. Dark regions correspond to features physically zeroed out 
in latent space ($m_i \approx 0$), confirming that the classifier made 
predictions without access to those visual cues.

\subsection{Failure Modes of ERSM}

\begin{figure}[h!]
    \centering
    \begin{minipage}{\linewidth}
        \centering
        (a) Focused Failure (Correct Mask, Wrong Class)
        \includegraphics[width=8cm]{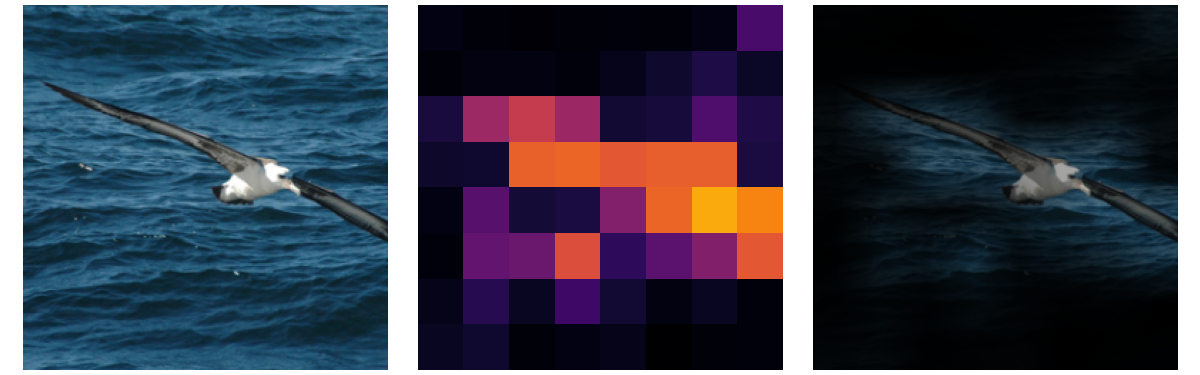}
    \end{minipage}
    \vspace{2mm}
    \begin{minipage}{\linewidth}
        \centering
        (b) Distracted Failure (Poor Mask, Background Noise)
        \includegraphics[width=8cm]{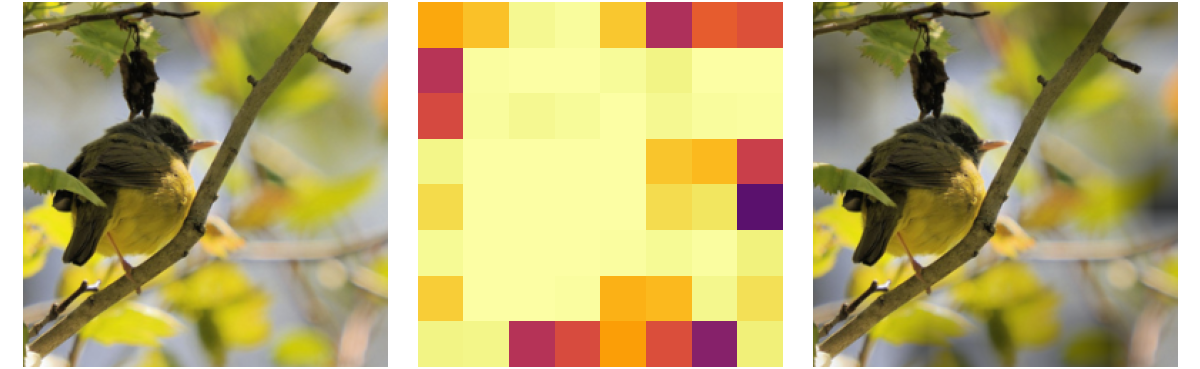}
    \end{minipage}
    \caption{Failure Mode. (a) The model correctly finds the bird but misclassifies it (fine-grained error). (b) The model fails to mask the background, leading to confusion.}
    \label{fig:failures}
\end{figure}

We categorize ERSM’s failure modes into two types (Figure~\ref{fig:failures}): \textbf{(a) Correct masking, misclassification.} The energy mechanism 
isolates the object and suppresses background, but the classifier fails 
to distinguish fine-grained traits. This highlights a limitation of the 
frozen backbone’s expressivity rather than the masking policy. {\bf (b) Masking failure.} The model retains substantial background 
because the energy potential cannot clearly separate object from environment. 
Such failures often occur in low-contrast images or when background textures 
resemble the object, confusing the classifier.

    

\section{Conclusion}
We introduced Energy-Regularized Spatial Masking (ERSM), a lightweight 
plug-and-play module that frames spatial feature selection as 
differentiable energy minimization. By combining unary and pairwise 
terms, ERSM produces input-adaptive, contiguous masks without 
segmentation supervision. Experiments show it preserves accuracy while 
improving interpretability and robustness: energy-guided patch removal 
outperforms random deletion and can even enhance performance by discarding 
distracting features. ERSM provides a simple, generalizable framework for 
energy-based feature selection in vision models.

\section*{Acknowledgements}

This work has received funding from the European Union Horizon Europe research and innovation programme under the Marie Sklodowska-Curie Actions (MSCA) grant agreement No. 101236749

\bibliographystyle{named}
\bibliography{ijcai26} 

\end{document}